\documentclass[lettersize,journal]{IEEEtran}
\usepackage{amsmath,amsfonts}
\usepackage{array}
\usepackage[caption=false,font=normalsize,labelfont=sf,textfont=sf]{subfig}
\usepackage{textcomp}
\usepackage{stfloats}
\usepackage{url}
\usepackage{verbatim}
\usepackage{multirow}
\usepackage{booktabs}
\usepackage{float}
\usepackage{graphicx}
\usepackage{algorithm}
\usepackage{algorithmic}

\def\BibTeX{{\rm B\kern-.05em{\sc i\kern-.025em b}\kern-.08em
    T\kern-.1667em\lower.7ex\hbox{E}\kern-.125emX}}
\usepackage{balance}
\begin{document}
\title{Compositional Alignment in Vision-Language Models}
\author{
\IEEEauthorblockN{
Ali Abdollahi\textsuperscript{1}\textsuperscript{\textsection}, 
Amirmohammad Izadi\textsuperscript{1}\textsuperscript{\textsection}, 
Armin Saghafian\textsuperscript{1},
Reza Vahidimajd\textsuperscript{1},
Mohammad Mozafari\textsuperscript{1},
Amirreza Mirzaei,
Mohammadmahdi Samiei\textsuperscript{1},
Mahdieh Soleymani Baghshah\textsuperscript{1}}

\IEEEauthorblockA{\textsuperscript{1}Department of Computer Engineering, Sharif University of Technology}

% \thanks{Manuscript created October, 2020; This work was developed by the IEEE Publication Technology Department. This work is distributed under the \LaTeX \ Project Public License (LPPL) ( http://www.latex-project.org/ ) version 1.3. A copy of the LPPL, version 1.3, is included in the base \LaTeX \ documentation of all distributions of \LaTeX \ released 2003/12/01 or later. The opinions expressed here are entirely that of the author. No warranty is expressed or implied. User assumes all risk.}
\thanks{This work has been submitted to the IEEE for possible publication. Copyright may be transferred without notice, after which this version may no longer be accessible.}
}

\markboth{Preprint}%
{How to Use the IEEEtran \LaTeX \ Templates}

\maketitle

\begingroup\renewcommand\thefootnote{\textsection}
\footnotetext{Equal contribution}
\endgroup

\begin{abstract}
Vision-language models (VLMs) like CLIP have showcased a remarkable ability to extract transferable features for downstream tasks. Nonetheless, the training process of these models is usually based on a coarse-grained contrastive loss between the global embedding of images and texts which may lose the compositional structure of these modalities. Many recent studies have shown VLMs lack compositional understandings like attribute binding and identifying object relationships. Although some recent methods have tried to achieve finer-level alignments, they either are not based on extracting meaningful components of proper granularity or don't properly utilize the modalities' correspondence (especially in image-text pairs with more ingredients). 
Addressing these limitations, we introduce \textbf{Com}positional \textbf{Align}ment (\textbf{ComAlign}), a fine-grained approach to discover more exact correspondence of text and image components using only the weak supervision in the form of image-text pairs. Our methodology emphasizes that the compositional structure (including entities and relations) extracted from the text modality must also be retained in the image modality. 
% We argue that VLMs already can provide proper embeddings for objects (while not necessarily for compositional scenes). 
To enforce correspondence of fine-grained concepts in image and text modalities, we train a lightweight network lying on top of existing visual and language encoders using a small dataset.  The network is trained to align nodes and edges of the structure across the modalities. Experimental results on various VLMs and datasets demonstrate significant improvements in retrieval and compositional benchmarks, affirming the effectiveness of our plugin model.
\end{abstract}

\begin{IEEEkeywords}
Class, IEEEtran, \LaTeX, paper, style, template, typesetting.
\end{IEEEkeywords}

\section{Introduction}
\label{Introduction}
\IEEEPARstart{V}{ision-Language} Models
(VLMs) have achieved impressive results in a broad range of vision-language tasks \cite{tan2019lxmert,bugliarello2021multimodal,radford2021learning,li2021align,zeng2021multi,gan2022vision}. The popular VLMs like CLIP \cite{radford2021learning}, and ALIGN \cite{jia2021scaling} focus on extracting global representation of images and texts by image and text encoders which are trained using a coarse-grained contrastive loss. 
Recent investigations have revealed that these VLMs struggle to comprehend compositional structures \cite{yuksekgonul2022and,thrush2022winoground,ma2023crepe}, such as binding attributes to the corresponding objects or identifying relationships between subjects and objects. ‌

To provide fine-grained VLMs, some models, such as PEVL \cite{yao2022pevl} and X-VLM \cite{pmlr-v162-zeng22c}, use more supervised datasets. In particular, they require fine-grained supervision, such as bounding box coordinates corresponding to a given entity. On the other hand, fine-grained VLMs like FILIP \cite{yao2021filip} don't need more supervision than image-text pairs. In these models, fine-grained similarities between regions of the image and words of the text are extracted and aggregated to get the overall similarity used in the usual contrastive learning.
PyramidCLIP \cite{gao2022pyramidclip} aligns image regions and object boxes with descriptive text. This model considers the local and global views for both the image and the text modalities and utilizes both Peer-level and Cross-level Alignment to tackle the mismatch of these modalities.

Despite introducing several fine-grained VLMs, these models don't properly utilize the correspondence of image and text modalities. For example, FILIP \cite{yao2021filip} proposes a simple way to create fine-grained supervision by dividing an image into patches and the descriptive text into tokens. This method considers each word of the text and each patch of the image as an independent component. For example, considering the phrase "A red flower", the "red" and "flower" tokens can be mistakenly matched to disjoint sets of patches without any losses. 

\begin{figure*}[ht!]
    \centering
    \label{fig:graphical}
    \includegraphics[width=0.95\textwidth]{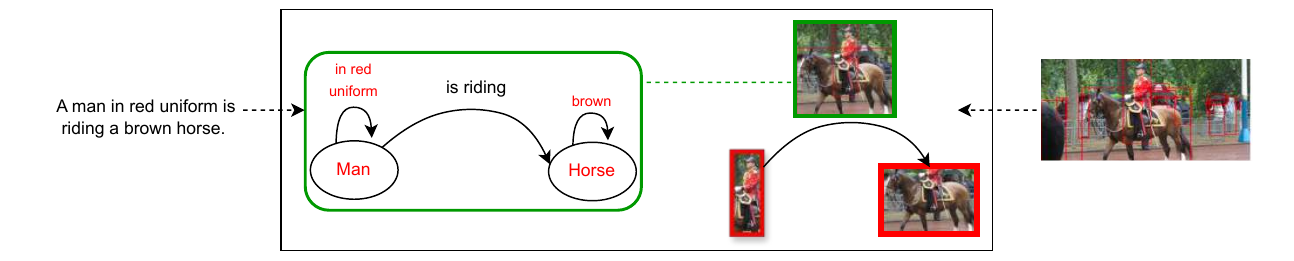} 
    \caption{Illustration of how entities and their relationships are considered components of the image and text.
    The textual modality contains entities and relationships shown as nodes and edges (i.e., actions) along with their two corresponding nodes (i.e., subject and object), respectively. The visual modality also mirrors this to provide a structure for better alignment of the modalities.}
\end{figure*}

To capture the correspondence of the text and image, the meaningful components of these modalities must be extracted. In the textual modality, the 
 Entity Relationship (ER) is utilized as a high-level conceptual model. An entity is a word indicating an object, such as "flower", and phrases like "red flower," which describes both the object and its attribute. Relations such as "a man riding a horse" correspond to a triplet that contains two entities (i.e., subject and object) and the specified relation between them. To provide a basis for better alignment of text and image, we also extract components of similar granularity for the visual modality by considering object-bounding boxes as candidate regions for visual entities and boxes including a pair of object-bounding boxes as candidate regions for visual relations \cite{johnson2015image}. Since entities and their attributes appear in the same area of an image in the visual modality, we consider both entities and described entities (with their attributes) as textual entity components. Therefore, the phrase "a red flower" as a textual entity must be aligned with the specific region of the image containing a red flower, even if the image also includes flowers of other colors.
A graph consisting of entities as its nodes and relationships as its edges can be used to denote textual components, as shown in Figure \ref{fig:graphical}.
The VLM can then be trained to align the compositional structures of the two modalities.

In this paper, we propose a method that efficiently utilizes a base VLM and provides a fine-grained VLM. Our method assumes that VLMs like CLIP can extract initial representations for entities of the text and objects of the image 
and need to be empowered by a lightweight model that can align the structure of the visual and textual modalities. Therefore, after extracting entities and relations from the text and identifying candidate regions for entities and relations from the image, the initial representation of these components is obtained using coarse-grained VLMs like CLIP.
To capture the compositional structure, ComAlign is trained on top of the frozen image and text encoders to provide fine-grained alignment of the image and text components.
This is done by modeling the compositional structures of the modalities as graphs and using a fine-grained matching strategy. This approach significantly improves zero-shot retrieval and compositional benchmark performance of base models while using a lightweight network and minimal training data. For example, it enhances the I2T retrieval performance of CLIP-ViT-B32 on MSCOCO\cite{lin2014microsoft} by 5.60\% and T2I by 6.27\%, surpassing PyramidCLIP, which employs the same backbone

The primary contributions of our work are outlined as follows:

\begin{enumerate}
    \item We designed a simple process to extract meaningful components from raw data.
    \item We implemented a straightforward strategy for unsupervised component matching and trained a lightweight network on top of the base VLM, enabling the alignment of VLM representations into fine-grained features.
    \item We enhanced performance across various benchmarks and different VLMs by utilizing a minimal dataset and avoiding the need to retrain the entire VLM.
\end{enumerate}

\section{Related Works}
\label{Related Works}

\subsection{Vision-Language Pretraining}

Vision-language pretraining aims to develop a unified embedding space that bridges the vision and language modalities by leveraging large-scale image-text sets.

In vision-language pretraining (VLP),  transformers are employed in two primary architectures. The first approach, single-stream models, integrates both the vision and language components into a single transformer. The second approach, called dual-stream models, utilizes separate transformer encoders for each modality, one for vision and one for language.
Prominent examples of single-stream models include UNITER \cite{chen2020uniter}, VisualBERT \cite{li1908visualbert}, VL-BERT \cite{su2019vl}, VILLA, Oscar \cite{li2020oscar}, and UNIMO \cite{li-etal-2021-unimo}. In contrast, dual-stream models are exemplified by CLIP \cite{radford2021learning}, ALIGN \cite{jia2021scaling}, ViLBERT \cite{lu2019vilbert}, DeCLIP \cite{li2021supervision}, COCA \cite{yu2022coca}, LXMERT \cite{tan2019lxmert}, and ALBEF \cite{li2021align}.

A common strategy in VLP involves using a masking technique on either the language or vision modality—or both—followed by reconstructing or classifying the masked elements to predict the omitted content. This technique is prevalent in models such as VisualBERT, LXMERT, Oscar, ViLBERT, ALBEF, VILLA, UNIMO, and UNITER. Another frequently used objective is image-text matching, which entails a binary classification task to determine whether an image and text pair are aligned. This objective underpins the pretraining of models like VisualBERT, VILLA, UNITER, ViLBERT, and ALBEF. However, these objectives often fall short of directly supporting the alignment of embeddings from matched pairs across the two modalities.

To address the alignment challenges inherent in these objectives, contrastive loss has been introduced in models like CLIP, ALIGN, COCA, UNIMO, ALBEF, and DeCLIP. This loss function enhances the alignment between the embeddings of the two modalities, though it can sometimes result in the neglect of fine-grained components, potentially leading to incorrect alignments.

In this paper, we tackle this limitation by utilizing dual-stream models pre-trained with contrastive learning using a frozen backbone. We then refine the alignment using a specialized fine-grained contrastive learning technique, thereby overcoming the previously mentioned weaknesses in cross-modal embedding alignment.

\subsection{Fine-grained Understanding}
When aligning coarse-grained embeddings across two modalities, mismatches can occur when elements present in one modality lack a corresponding element in the other, leading to potential false alignments. Additionally, this approach may overlook the finer details within each modality, failing to align them accurately with their counterparts. Some methods have introduced multi-level semantic components to address these issues, as seen in models like OSCAR, VinVL, MVPTR, X-VLM, and PyramidCLIP.

In OSCAR and VinVL, the focus is on the visual modality, where images are broken down into object boxes and their associated tags. VinVL builds on OSCAR by pre-training a more advanced object-attribute detector, improving performance.

MVPTR takes a more comprehensive approach by constructing two levels of semantic components for both visual and linguistic modalities. In the visual modality, images are decomposed into object boxes with position-aware features, while object tags serve as inputs. For the linguistic modality, the model processes the token sequence of the text and also incorporates phrase-level inputs derived from a scene graph extracted from the main text. On the other hand, X-VLM identifies visual concepts based on extracted phrases and aligns them with visual components at various levels of granularity. 
However, these methods do not directly match each fine-grained component representation with its corresponding counterpart in the other modality, as done in contrastive learning, and directly match coarse-grained representations. Our method, on the other hand, directly aligns each fine-grained component representation with its corresponding one in the other modality without any extra alignment supervision between components of both modalities. As a result, in our embedding space, the representations of fine-grained components are closer to those of the corresponding components in the other modality.

% However, these methods often struggle to achieve accurate alignment between components within each modality and their counterparts in the other modality.

Our method addresses these shortcomings by not only extracting multi-level semantic components more effectively but also ensuring accurate alignment between each component and its corresponding counterpart across modalities.

Various image-text matching methods have been developed to improve alignment accuracy between images and textual components. For instance, SCAN \cite{lee2018stacked} and its variants \cite{diao2021similarity,liu2019focus,wu2019learning} employ a cross-attention mechanism to assess the relevance of each component from one modality to all components of the other modality. NAAF \cite{zhang2022negative} introduces a novel approach by calculating the negative similarity of mismatched components alongside the positive similarity of matched pairs. CHAN \cite{pan2023fine}, on the other hand, uses hard matching of text and image components, presuming that each textual entity corresponds to a specific region in the image, though the reverse is not necessarily true. This method utilizes max pooling over image regions for each textual entity within their similarity matrix. FILIP \cite{yao2021filip} takes a bidirectional approach, assuming that each text component corresponds to an image component and vice versa, performing average pooling over image components for each textual component and vice versa.

% Despite these image-text matching methods advancements, they often fall short in accurately creating multi-level fine-grained components within each modality. As a result, while they achieve alignment between corresponding components across modalities, the quality of these aligned components is often inferior compared to our proposed method. 

Despite advancements in these image-text matching methods, their composed components are not broken down into distinct semantic categories as our method does. Our approach uses a hard assignment method similar to FILIP but extends it by extracting three levels of semantic components within each modality. This ensures that components at each level are matched precisely to their corresponding level in the other modality, resulting in more accurate and meaningful cross-modal alignments.

\section{Proposed Method}
\label{method}

\begin{figure*}
  \centering
  \label{Matching Method}
  % \fbox{\rule[-.5cm]{0cm}{4cm} \rule[-.5cm]{4cm}{0cm}}
  \includegraphics[width=\linewidth]{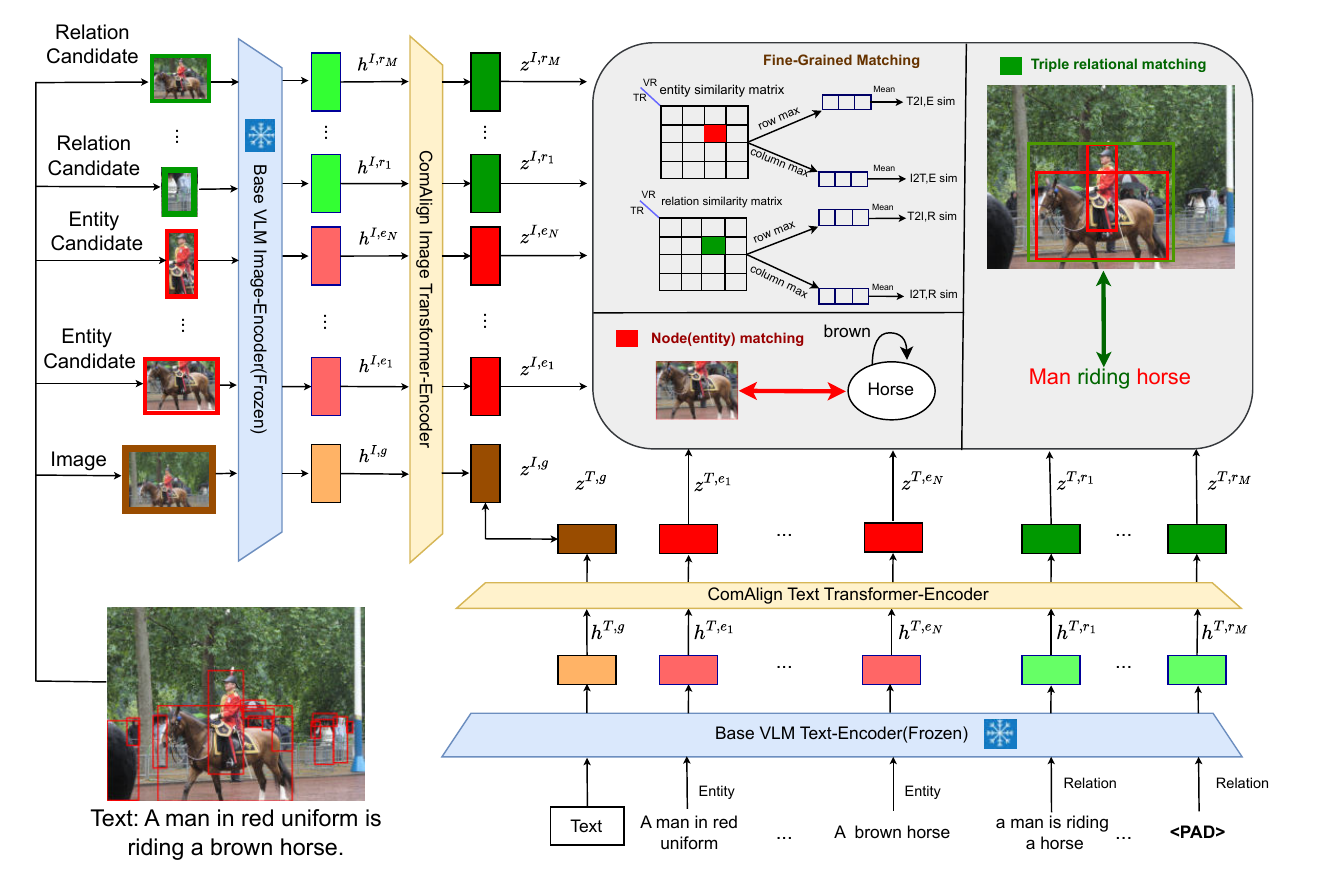}
  \caption{Overview of the proposed method. Given a batch of image-text pairs, each image and text is preprocessed by object-detector and NLP tools to extract entity and relational components. These components, along with the original image and text, are processed by a base VLM to obtain visual and textual representations. These are then passed through our ComAlign image and text encoders.
  We calculate the similarity score between an image and a text using three metrics:
  1) Coarse-grained similarity: Calculated as the dot product of the global features of the image and text.
  2) Fine-grained entity-based similarities: The entity similarity matrix is obtained by calculating the cosine similarity between each pair of the visual entity representation (VR) and textual entity representation (TR). 
  3) Fine-grained relation-based similarities: Similarly, the relation similarity matrix is computed according to the cosine similarity of all pairs of visual and textual relation representations.
  By employing Fine-Grained Matching on the obtained matrices, the whole entity-based similarity and relation-based similarity between the image and text are found (for both Text2Image and Image2Text directions). These fine-grained similarities are used in the contrastive training and inference process.
  } 
\end{figure*}

In this section, we explain our proposed method for extracting and aligning the compositional structure of image and text. Initially, we extract fine-grained components from images and texts. These components, along with the entire image and text, are processed by a frozen pre-trained VLM to obtain representations. We then feed them into our ComAlign encoders to capture the interactions between the fine-grained and coarse-grained features within each modality.
By aligning corresponding concepts across modalities, we achieve representations that effectively capture both fine-grained and coarse-grained information.
% Matching each category of components with corresponding components in the other modality yields aligned embeddings that capture fine-grained and coarse-grained information.
% Our method enhances compositional understanding and fine-grained details within both modalities while maintaining coarse-grained features.

% In this section, we propose our method \textit{Model Name} for better fine-grained and coarse-grained image and text alignment. Initially, we employ an object detection model to extract components from images, supplemented by spaCy \cite{spaCy} for extracting fine-grained elements from the text. These elements form the basis for generating captions using the VLM template. Captions and images are then embedded using the base VLM. We concatenate the VLM coarse-grained embeddings to previous fine-grained embeddings and pass them through a lightweight network (two transformer layers) to obtain fine-grained embeddings. The goal is to extract fine-grained features from the earlier embeddings of VLMs that were not previously used by aligning the base models' embedding. This is achieved by utilizing \textit{Fine-Grained Loss}, while coarse-grained features contribute to \textit{Contrastive Loss} calculations. 

\subsection{Extraction of Fine-grained Components}
\label{subsec:preprocess}

The structured nature of textual modality allows us to extract entity and relational components from text more accurately than images. However, since images lack this inherent structure, we utilize an object detector to extract candidate entities and relations. More precisely, we present a preprocessing method for extracting fine-grained components from existing image and text data that initially lacked detailed annotations. This process focuses on obtaining two distinct types of fine-grained components: \textit{1) Entity Components}, which are extracted to represent individual objects within an image or to identify a noun (along with its corresponding adjective, if any) within the text. \textit{2) Relational Components} are designed to capture the interactions between two objects within the image or the connections between two entities linked by a specific relationship in the text.
We utilize SpaCy's "en\_core\_web\_sm" pre-trained English language model \cite{spaCy} to extract nouns and their corresponding adjectives as textual entity components and $r=(subject, action, object)$ triplets as textual relational components.

For the images, we employ an object detector to extract object-bounding boxes within images as visual entity components. Moreover, we consider all possible pairs of object-bounding boxes identified in the previous step to extract candidate regions for relations. We crop a minimal bounding box for each pair of objects containing both entities as a candidate relational component. Our object detector provides a confidence score for each detected object, reflecting its likelihood of objectness. For each relation candidate, we calculate a score by multiplying these confidence scores. The candidates with the highest scores are chosen as the final candidates of relational components.

% These elements are directly extracted from the provided data in datasets that include fine-grained annotations. We generate fine-grained expressions from these elements using a predefined template.

\subsection{Architecture}
\label{Architecture}

First, we embed the extracted textual and visual components. More precisely, the object and relation bounding boxes in the image are cropped, resized, and then embedded by the image encoder of the base VLM. Textual entities and relations are also embedded by the text encoder of the base VLM. Moreover, the frozen VLM also embeds the whole image and text. The obtained representations for the $N$ entity components, the $M$ relational components, and the global representations of image $i$ are shown as $\{h_i^{I,e}\}_{e=1}^N$,
$\{h_i^{I,r}\}_{r=1}^M$, and $h_i^{I,g}$, respectively. The corresponding representations for the entities, relations, and the whole input for text $j$ are also denoted as $\{h_j^{T,e}\}_{e=1}^N$, $\{h_j^{T,r}\}_{r=1}^M$, and $h_j^{T,g}$, respectively. $N$ and $M$ are treated as hyper-parameters, determined before training. The number of extracted entities and relationships is adjusted by truncating excess components or padding to reach the determined numbers $N$ and $M$, respectively.

We want to improve the representations of fine-grained components since they have been extracted individually by the base VLM. To this end, we employ a simple two-layer transformer architecture to find the contextualized representations of components that also have been enforced to consider the fine-grained and coarse-grained correspondence of the image and text modality. 
Therefore, the representation of image $i$ and text $j$ are fed as $h_i^I=[h_i^{I,g},h_i^{I,e_1}, ...,h_i^{I,e_N},h_i^{I,r_1},...,h_i^{I,r_M}]$ and $h_j^T=[h_j^{T,g},h_j^{T,e_1}, ...,h_j^{T,e_N},h_j^{T,r_1},...,h_j^{T,r_M}]$ to the ComAlign image and text encoder, respectively.  Specifically, the contextualized representations are obtained as:

\begin{equation}
    \label{eq:eq1}
    z_i^I =
    F_{\theta_I}(h_i^I),
    \quad  \quad 
    z_j^T =
    G_{\theta_T}(h_j^T),
\end{equation}

Where the ComAlign encoder networks $F_{\theta_I}$ and $G_{\theta_T}$ are two-layer transformer models for improving vision and language representations, respectively. %As depicted in Figure \ref{Matching Method}, the image feature vector $z^I$ can be decomposed into three components: $\{z^{I, e}\}_{e=1}^N\in \mathbb{R}^{ N \times D}$, $\{z^{I, r}\}_{r=1}^M\in \mathbb{R}^{ M \times D}$ , and $z^{I, g} \in \mathbb{R}^{ 1 \times D}$. Here, $z^{I, g}$ represents the coarse-grained image embedding, $z^{I, e}$ denotes the fine-grained embeddings of N entity components within the image, and $z^{I, r}$ captures the fine-grained embeddings of $M$ relational components. Similarly, $z^T$ is composed of similar components, $z^{T, e}$, $z^{T, r}$, and $z^{T, g}$.

\subsection{Training Objectives}
\label{objective}

% As outlined in Section \ref{Architecture}, our approach involves training a lightweight, two-layer transformer on the encoders for each modality, denoted as $F_{\theta_I}$ for images and $G_{\theta_T}$ for texts. 
The goal is to ensure that each image's representation closely aligns with its corresponding text while simultaneously differing significantly from the representations of unrelated texts. To achieve this, we must match the corresponding components in the image and text pair. First, we define the fine-grained matching method for aligning image and text representations. Then, we indicate how entity, relational, and global similarity between image and text representations are obtained.

\textbf{Fine-Grained Matching} We intend to match the corresponding components of two modalities. To do this, we use the matching strategy introduced in FILIP \cite{yao2021filip}
%Assuming $x \in \mathbb{R}^{N \times D}$ and $x' \in \mathbb{R}^{M \times D}$ are the components in both modalities, then the similarity of each image and text pair is computed as follows:
and align the two set of representation vectors $\{x_k\}_{k=1}^{C}$ and $\{x'_k\}_{l=1}^{C'}$, using the following Fine-Grained-Matching (FGM) function:
\begin{align}
\begin{split}
    &FGM(\{x_k\}_{k=1}^{C}, \{x'_l\}_{l=1}^{C'}) = \\
&\text{mean}_{1 \leq k \leq C} \left \{ \text{max}_{1 \leq l \leq C'} \left \{ x_k^T x'_l \right \} \right \}.
\end{split}
\end{align}

\textbf{Entity and Relational Components Similarity} We compute the entity-based similarity between images and text by defined fine-grained matching. Image-to-Text (I2T) and Text-to-Image (T2I) similarities of image $i$ and text $j$ is defined as follows:

\begin{align}
\label{eq:ent_sim}
\begin{split}
&s_{i,j}^{I2T , E} = FGM(\{z_i^{I, e}\}_{e=1}^{N},
\{z_j^{T, e}\}_{e=1}^{N}),\\[0.5em]
&s_{i,j}^{T2I, E} = FGM(\{z_j^{T, e}\}_{e=1}^{N},
\{z_i^{I, e}\}_{e=1}^{N}),
\end{split}
\end{align}

where $z_i^{I, e} \in \mathbb{R}^{D}$ and  
$z_j^{T, e} \in \mathbb{R}^{D}$ shows the representation of the entity component $e$ of image $i$ and text $j$ respectively, and $N$ denotes the number of entity components.

Relational components are matched similarly:

\begin{equation}
\label{eq:rel_sim}
\begin{split}
&s_{i,j}^{I2T, R} = FGM(\{z_i^{I, r}\}_{r=1}^{M}, \{z_j^{T, r}\}_{r=1}^{M}) , \\[0.5em]
&s_{i,j}^{T2I, R} = FGM(\{z_j^{T, r}\}_{r=1}^{M}, \{z_i^{I, r}\}_{r=1}^{M} ) ,
\end{split}
\end{equation}
where $z_i^{I, r} \in \mathbb{R}^{D}$ and  
$z_j^{T, r} \in \mathbb{R}^{D}$ are relational representations of image $i$ and text $j$, and $M$ is the number of relational components.

\textbf{Global Similarity}
We use the standard dot product for computing the similarity between two global features, considering $z_i^{I, g} \in \mathbb{R}^{D}$ and $z_i^{T, g} \in \mathbb{R}^{D}$:

\begin{equation}
\label{eq:glob_sim}
s_{i, j}^{I2T, G} = s_{i, j}^{T2I, G} = (z_i^{I, g})^T z_j^{T, g}.
\end{equation}

The loss function is the sum of the contrastive losses for each of the entity, relational, and global features, with the similarity calculated differently for each category. Specifically, the image-to-text and text-to-image contrastive losses are defined as:

\begin{equation}
\label{eq:6}
\fontsize{9}{10}\selectfont
\begin{aligned}
\mathcal{L}_i^{I2T} = 
& f_i(\{s_{i, j}^{I2T, E}\}_{j=1}^B) 
+ f_i(\{s_{i, j}^{I2T, R}\}_{j=1}^B) 
+ f_i(\{s_{i, j}^{I2T, G}\}_{j=1}^B) , 
\end{aligned}
\end{equation}

\begin{equation}
\label{eq:7}
\fontsize{9}{10}\selectfont
\begin{aligned}
\mathcal{L}_i^{T2I} = 
& f_i(\{s_{i, j}^{T2I, E}\}_{j=1}^B) 
+ f_i(\{s_{i, j}^{T2I, R}\}_{j=1}^B) 
+ f_i(\{s_{i, j}^{T2I, G}\}_{j=1}^B),
\end{aligned}
\end{equation}

\noindent where $f_i$ is defined as:

\begin{equation}
    f_i(\{s_{i,j}\}_{j=1}^B) = - \log \frac{\exp(s_{i, i})}{\sum_{j=1}^{B} \exp(s_{i, j})}.
\end{equation}

\noindent The final loss in a batch is computed by mean of I2T and T2I losses as:

\begin{equation}
\mathcal{L} = \frac{1}{2B} \sum_{i=1}^{B} (\mathcal{L}_i^{I2T} + \mathcal{L}_i^{T2I}).
\end{equation}

\begin{figure}[h]
    \centering
    \includegraphics[width=\linewidth]{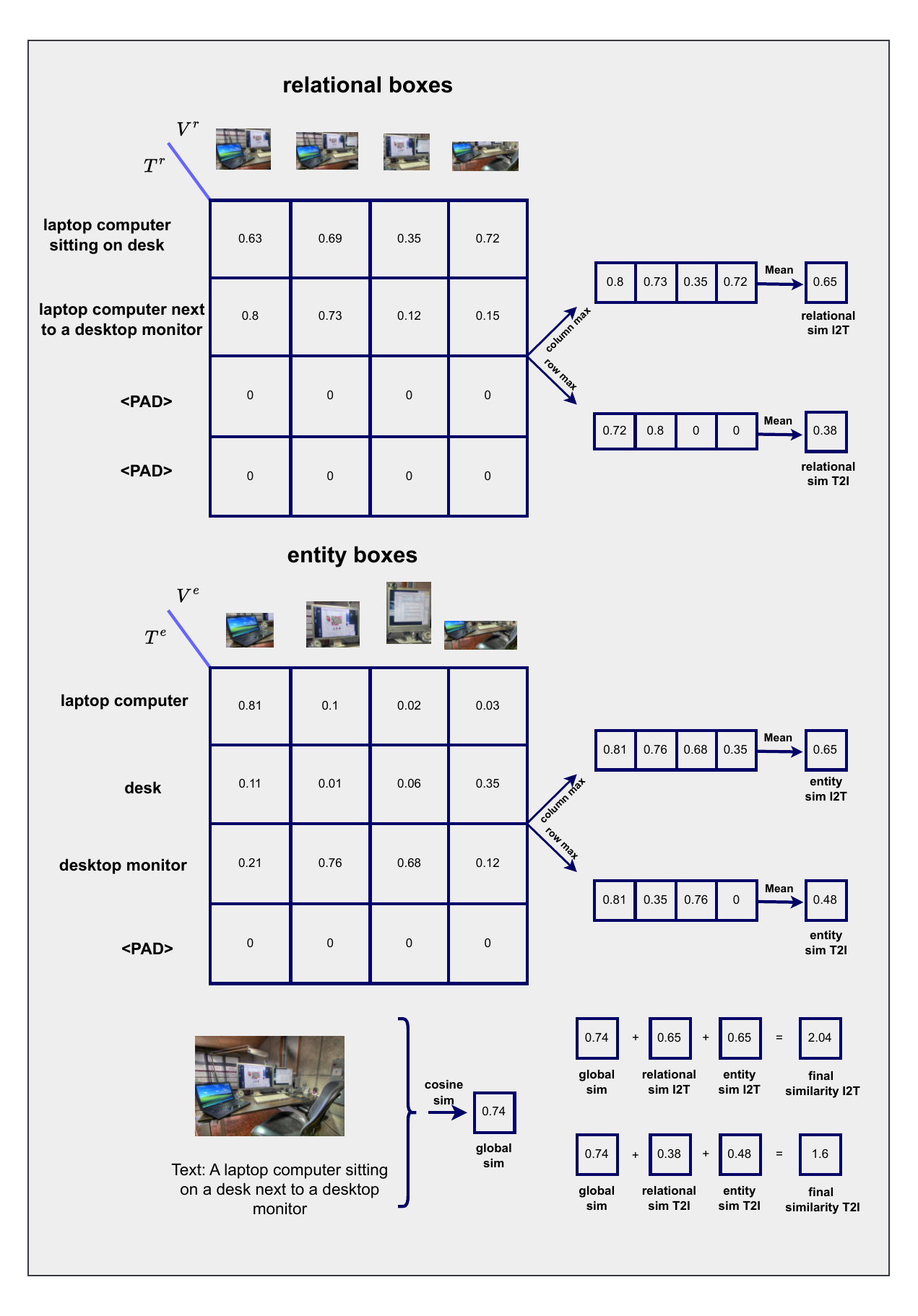}
    \caption{Illustration of the process of calculating Image-to-Text (I2T) and Text-to-Image (T2I) similarity, including global, entity, and relational components.}
    \label{similarity-example}
\end{figure}

\noindent
Figure \ref{similarity-example} shows an example of this calculation process.

\subsection{Inference}
\label{subsec:infer}
%During inference, entity and relational components of images and texts are extracted using the same method as section \ref{subsec: preprocess}. We then utilize the base VLM's encoder to embed these features as well as the original image and text, resulting in three feature sets for each modality $h_i^{I, e}$, $h_i^{I, r}$, $h_i^{I, g}$. These features are then concatenated and passed through our network to obtain the aligned embeddings $z_i^{I, e}$, $z_i^{I, r}$, $z_i^{I, g}$, Text embeddings are calculated similarly.
During inference, the fine-grained and coarse-grained representation of the images and texts ($z^I, z^T$) are obtained according to the proposed method in Section \ref{Architecture}.
To calculate the T2I and I2T similarities between each image-text pair, we consider a weighted sum of corresponding fine and coarse-grained similarities ($s^{.,E}, s^{.,R}, s^{.,G}$) along with the dot product of the base VLM image and text representations ($h^{I,g}, h^{T,g}$).

Additionally, we use fine-grained T2I similarities to calculate the final I2T similarity score. 
% This approach is based on the premise that every component in a text exists in its corresponding image, but not all details of an image are described in its caption. Therefore, incorporating T2I fine-grained similarities helps capture components that I2T similarities might miss.
This approach is based on the premise that I2T similarity cannot capture all the components present in the image because not all details of an image are described in its caption. Therefore, incorporating T2I fine-grained similarities could help compensate for this weakness.
The final similarity score is formulated as follows:

\begin{align}
% \fontsize{10}{10}\selectfont
\begin{split}
s_{i, j}^{I2T} = &(h_i^{I, g})^T h_j^{T, g} \\
&+ \alpha_1 (s_{i, j}^{I2T, G} + s_{i, j}^{I2T, E} +  s_{i, j}^{I2T, R}) \\
&+  \alpha_2  (s_{i, j}^{T2I, E} + s_{i, j}^{T2I, R} ),
\label{eq:10}
\end{split}
\end{align}

\begin{equation}
    s_{i, j}^{T2I} = (h_i^{I, g})^T h_j^{T, g} +
    \beta_1 ( s_{i, j}^{T2I, G} + 
    s_{i, j}^{T2I, E} + s_{i, j}^{T2I, R}).
    \label{eq:11}
\end{equation}

\noindent
Here $s_{i, j}^{I2T/T2I, E/R/G}$ are calculated according to Equations \ref{eq:ent_sim}, \ref{eq:rel_sim}, and \ref{eq:glob_sim}. Also, $\alpha_1$, $\alpha_2$ and $\beta_1$ are considered as hyper-parameters.

\section{Experiments}
\label{Experiments}

\subsection{Experimental Setup}

\textbf{Base VLMs}
We applied our alignment method to two CLIP model backbones released by OpenAI: ViT-B/32 and ViT-L/14.
Furthermore, we tested our method on two other models: NegClip \cite{yuksekgonul2022and} and COCA \cite{yu2022coca}. NegCLIP leverages negative samples to improve contrastive learning, enhancing the model's ability to distinguish between similar images and texts. 
Meanwhile, COCA employs a caption generation objective in addition to contrastive learning, which helps improve fine-grained understanding.

We used the NegCLIP checkpoint from the official GitHub repository
\footnote{\url{https://github.com/mertyg/vision-language-models-are-bows}},
and the COCA  model using a checkpoint from OpenCLIP \cite{ilharco_gabriel_2021_5143773}, which was trained on the LAION-2b dataset \cite{schuhmann2022laionb}. Both models use the ViT-B/32 network as their backbone.

% Most VLMs compare their performance with the CLIP-ViT-L/14 model. However, to fairly compare our method with PyramidCLIP (which uses a ViT-B/32 backbone and is trained on 143 million training samples), we applied our alignment method to the CLIP-ViT-B/32 model to ensure a consistent backbone with PyramidCLIP. In addition, we tested our alignment method on the COCA \cite{yu2022coca} model using a checkpoint from OpenCLIP \cite{ilharco_gabriel_2021_5143773}, which was trained on the LAION-2b dataset \cite{schuhmann2022laionb}. The COCA model, designed for generative tasks, enhances fine-grained understanding. Furthermore, for NegCLIP, we use the pre-trained checkpoint provided in the official repository.

\textbf{Implementation Details}
\label{implementation}
The training process was performed on a Nvidia 1080 GPU, with each base model training completed within 4 hours (All our experiments were also performed on the specified GPU card). This minimal training time and low GPU VRAM requirement are due to our lightweight network, which consists of a two-layer transformer and a relatively small training dataset of approximately 80,000 image-text pairs. 
All models were trained using the AdamW optimizer \cite{loshchilov2017decoupled} and a StepLR learning rate scheduler, with a batch size of 1600 image-text pairs. Positional encoding was implemented in the transformer layers as described in \cite{NIPS2017_3f5ee243}. 
We set the maximum number of both entity and relational components to 10. 
We used spaCy to extract components from the text, and YOLOv9 \cite{wang2024yolov9} as the object detector to identify visual components. The training process is depicted in Algorithm \ref{alg}.

We performed hyper-parameter tuning for training and inference using a subset of the training dataset we composed. Further details can be found in Appendix \ref{app:Hyper-parameter Tuning}.

\begin{algorithm}[H]
\caption{Training process of ComAlign}
\label{alg}
\begin{algorithmic}[1]

% \STATE Initialize preprocessed dataset $\mathcal{D}$, batch size $\mathcal{B}$
% \STATE Initialize transformers $F_{\theta_I}$ and  $G_{\theta_T}$
% \FOR{update step $=1,M$}
%     \STATE Sample image-text pair representations $(h^I_i, h^T_i)_{i=1}^{\mathcal{B}}$
%     \STATE Calculate $z_i^I = F_{\theta_I}(h_i^I)$ and $z_i^T = G_{\theta_T}(h_j^T)$
%     \STATE Separate each $z_i^I$ and $z_i^T$ into their entity, relational, and global representations ($z_i^{-,e}, z_i^{-,r}, z_i^{-,g}$)
%     \STATE Calculate the similarity score ($s^{-,E}, s^{-,R}, s^{-,G}$) between each image and text using Equations \ref{eq:ent_sim}, \ref{eq:rel_sim}, \ref{eq:glob_sim}
%     \STATE Calculate the final loss via Equations 6-8
%     \STATE Update the parameters of the transformers ($F_{\theta_I}$, $G_{\theta_T}$) by Adam optimizer;
    
% \ENDFOR

\STATE Initialize preprocessed dataset $\mathcal{D}$ and batch size $\mathcal{B}$
\STATE Initialize transformer models $F_{\theta_I}$ and $G_{\theta_T}$
\FOR{update step $= 1$ to $M$}
    \STATE Sample a batch of image-text pairs $(h^I_i, h^T_i)_{i=1}^{\mathcal{B}}$ from $\mathcal{D}$
    \STATE Compute image and text representations: $z_i^I = F_{\theta_I}(h_i^I)$ and $z_i^T = G_{\theta_T}(h_i^T)$
    \STATE Decompose $z_i^I$ and $z_i^T$ into entity, relational, and global features $z_i^{I,e}, z_i^{I,r}, z_i^{I,g}$ and $z_i^{T,e}, z_i^{T,r}, z_i^{T,g}$, respectively.
    \STATE Calculate similarity scores for entities, relations, and global features: $s^{.,E}$, $s^{.,R}$, $s^{.,G}$ using Equations \ref{eq:ent_sim}, \ref{eq:rel_sim}, and \ref{eq:glob_sim}
    \STATE Compute the final loss using Equations (6) through (9)
    \STATE Update model parameters $\theta_I$ and $\theta_T$ of $F_{\theta_I}$ and $G_{\theta_T}$ using the Adam optimizer
\ENDFOR

\end{algorithmic}
\end{algorithm}

\subsection{Datasets}
\textbf{Visual Genome} This dataset comprises 100,000 images with fine-grained annotations. Each image includes two types of annotations: 1) Attribute Annotations: These annotations describe the objects and their attributes.
2) Relational Annotations: These annotations consist of triplets in the format (Subject, Object, Relation).

\textbf{MSCOCO} This dataset contains approximately 100,000 images, each accompanied by five descriptive captions. We used the version of MSCOCO released in 2017, which consists of 118K images in the training split and 5K in the validation split.

\textbf{Flickr30K} This dataset includes around 30,000 images, each with several captions similar to MSCOCO.

\subsection{Zero-shot Image-Text Retrieval}
\label{subsec:retrieval}
Zero-shot image-text retrieval consists of two sub-tasks: image-to-text retrieval and text-to-image retrieval. We use the popular MSCOCO \cite{lin2014microsoft} and Flickr30K \cite{plummer2015flickr30k} datasets for both training and evaluations, along with the addition of Visual Genome \cite{yuke2017visual} dataset, explicitly used for training. 

We compare the performance of our alignment method on Flickr30K and MSCOCO datasets against the base VLMs and PyramidClip \cite{gao2022pyramidclip}. PyramidClip full-finetunes CLIP on a data set of 143M samples. It creates multiple semantic levels and performs contrastive alignment between them, which helps the model with better compositional understanding.

For the Flickr30K zero-shot retrieval, we trained our model on around 100K image-text pairs from the Visual Genome dataset, excluding images that are also part of Flickr30K. For the MSCOCO zero-shot retrieval, we removed images from the Visual Genome dataset present in MSCOCO alongside training data from Flickr30K, resulting in a dataset of approximately 80K image-text pairs.

Table \ref{tb:table1} shows our results compared to the baselines. We observe extensive performance improvement for CLIP-ViT-B/32 and CLIP-ViT-L/14 in image-to-text (I2T) and text-to-image (T2I) retrieval on both datasets. Additionally, our method improves COCA's performance, except in text-to-image retrieval on the Flickr30K dataset, where our performance was comparable to the base model. Notably, while NegCLIP employs a full-finetuning approach that improves compositional understanding using negative image-text pairs, our contribution complements theirs. Hence, when applying our method on NegCLIP, we achieved up to a 3.06\% improvement in image-to-text retrieval on Flickr30K.

Most interestingly, when applied to CLIP-ViT-B/32, our model outperforms PyramidCLIP in image-to-text retrieval. This is significant because, despite both models using identical backbones, PyramidCLIP finetunes the entire network with a massive dataset (143M samples). In contrast, we only train a small network on top of the base CLIP using a much smaller dataset (100K samples). We believe that our careful construction of entity and relation components, combined with a straightforward matching strategy, enables our method to utilize the fine-grained information in the base models effectively.

% with complex attention layers including Unicoder-VL (Li et al., 2020a), ImageBERT (Qi et al., 2020), UNITER (Chen et al., 2020), VILLA (Gan et al., 2020), ERNIE-ViL (Yu et al., 2021), Oscar (Li et al., 2020b), VinVL (Zhang et al., 2021), ALBEF (Li et al., 2021a), and methods trained on larger-scale image-text datasets including CLIP (Radford et al., 2021) and ALIGN (Jia et al., 2021).

% Our method achieves better results across all metrics on both Flickr30K and MSCOCO datasets, with improvements of up to 6.7\% R@1 in MSCOCO text-to-image retrieval, except for zero-shot text-to-image retrieval on Flickr30K, where we achieve comparable results.

% This enhancement likely stems from our specialized matching method, which aligns each extracted relational and descriptive expression with corresponding fine-grained relational and descriptive boxes. Additionally, we utilize the similarity between coarse-grained embeddings of images and text that is attended to fine-grained relational and descriptive compositions by passing through trained two-layer transformers of our extension. This approach ensures that coarse-grained embeddings are not discarded and are instead integrated with fine-grained embeddings to enhance relevance and accuracy.

\begin{table*}
  \caption{Zero-shot image-text retrieval results on MSCOCO and Flickr30K datasets.}
  \label{tb:table1}
  \centering
  \setlength{\tabcolsep}{4pt}
  \resizebox{1\textwidth}{!}{
  \begin{tabular}{lllllllllllll}
  \toprule
   \multicolumn{1}{c}{\multirow{4}{*}{\large \textbf{Method}}}& \multicolumn{6}{c}{\textbf{MSCOCO}} & \multicolumn{6}{c}{\textbf{Flickr30K}} \\
   \cmidrule(r){2-7} 
   \cmidrule(r){8-13}
   & \multicolumn{3}{c}{image-to-text} & \multicolumn{3}{c}{text-to-image} & \multicolumn{3}{c}{image-to-text} & \multicolumn{3}{c}{text-to-image} \\
   \cmidrule(r){2-4} 
   \cmidrule(r){5-7} 
   \cmidrule(r){8-10} 
   \cmidrule(r){11-13}
   & R@1 & R@5 & R@10 & R@1 & R@5 & R@10 & R@1 & R@5 & R@10 & R@1 & R@5 & R@10 \\ 
   \midrule
   CLIP-ViT-B/32            & 50    & 74.96 & 83.28 & 30.35 & 54.77 & 66.09 & 78.59 & 95.36 & 97.63 & 59.72 & 84.83 & 90.67 \\[-0.4ex]
   CLIP-ViT-B/32 + ComAlign & 55.60 & 79.72 & 86.88 & 36.62 & 63.55 & 74.77 & 82.24 & \textbf{97.04} & \textbf{98.61} & 66.27 & 88.22 & 93.11 \\[-0.4ex]
   Relative gain            & 5.60  & 4.76  & 3.60  & 6.27  & 8.78  & 8.68  & 3.65  & 1.68  & 0.98  & 6.55  & 3.39  & 2.44 \\
   % \midrule
   % CLIP-ViT-L/14            & 56.08 & 79.6  & 86.86 & 35.31 & 59.96 & 70.14 & 86.29 & 97.33 & 99.30 & 67.83 & 88.85 & 93.25 \\[-0.4ex]
   % CLIP-ViT-L/14 + ComAlign & 61.86 & 84.34 & 90.80 & 42.40 & 69.04 & 78.78 & 89.25 & 97.92 & 99.30 & 73.19 & 91.97 & 95.44 \\[-0.4ex]
   % Relative gain            & 5.78  & 4.74  & 3.94  & 7.09  & 9.08  & 8.64  & 2.96  & 0.59   & 0    & 5.36  & 3.12  & 2.19 \\
   \midrule
   COCA-ViT-B/32            & 54.04 & 77.72 & 86.08 & 35.89 & 61.20 & 71.97 & 82.64 & 95.36 & 97.63 & 64.31 & 86.96 & 91.77 \\[-0.4ex]
   COCA-ViT-B/32 + ComAlign & 56.42 & 80.30 & 88.06 & 37.29 & 63.98 & 74.93 & 84.22 & 96.64 & 98.32 & 63.07 & 86.31 & 92.05 \\[-0.4ex]
   Relative gain            & 2.38  & 2.58  & 1.98  & 1.40  & 2.78  & 2.96  & 1.58  & 1.28  & 0.69  & -1.24 & -0.65  & 0.28 \\
   \midrule
   NegClip-ViT-B/32            & 56.84 & 80.72 & 88.06 & 41.56 & 68.68 & 78.92 & 83.03 & 95.56 & 97.53 & 68.73 & 89.90 & 94 \\[-0.4ex]
   NegClip-ViT-B/32 + ComAlign & \textbf{58.60} & \textbf{82.62} & \textbf{89.42} & \textbf{42.16} & \textbf{69.82} & \textbf{79.93} & \textbf{86.09} & 96.74 & 98.22 & \textbf{69.11} & \textbf{90.43} & \textbf{94.49} \\[-0.4ex]
   Relative gain               & 1.76  & 1.90  & 1.36  & 0.60  & 1.14  & 1.01  & 3.06  & 1.18  & 0.69  & 0.38  & 0.53  & 0.49 \\
   \midrule

   PyramidCLIP-ViT-B/32 & 52.6     & 79.04     & 86.8   & 39.64  & 65.14 & 75.37 & 80.96 & 96.64     & \textbf{98.61}   & 67.31   & 89.30  & 93.53   \\ \midrule \midrule
    CLIP-ViT-L/14            & 56.08 & 79.6  & 86.86 & 35.31 & 59.96 & 70.14 & 86.29 & 97.33 & \textbf{99.30} & 67.83 & 88.85 & 93.25 \\ [-0.4ex]
   CLIP-ViT-L/14 + ComAlign & \textbf{61.86} & \textbf{84.34} & \textbf{90.80} & \textbf{42.40} & \textbf{69.04} & \textbf{78.78} & \textbf{89.25} & \textbf{97.92} & \textbf{99.30} & \textbf{73.19} & \textbf{91.97} & \textbf{95.44} \\[-0.4ex]
   Relative gain            & 5.78  & 4.74  & 3.94  & 7.09  & 9.08  & 8.64  & 2.96  & 0.59   & 0    & 5.36  & 3.12  & 2.19 \\ \bottomrule
   
    % PyramidCLIP-ViT-B/32 + Our  & 53.28      & 80.02        & 87.58                 & 40.18   & 66.30 & 76.64 & 82.44  & 97.33     & 98.61       & 67.21   & 89.44  & 93.86 \\[-0.5ex]
    % Relative Gain          & 0.68       & 0.98   & 0.78   & 0.54      & 1.16 & 1.27 & 1.48 & 0.69   & 0       & -0.1   & 0.14  & 0.33         \\ 

\end{tabular}
}
\end{table*}

% \subsection{Zero-shot Classification}
% In the following section, we evaluate our proposed method on the zero-shot image classification task using five downstream classification datasets. Table \ref{tb:table2} illustrates the results of our alignment method compared to the base VLMs across these datasets. The results of our zero-shot image classification experiments indicate that our method yields comparable performance to existing models across various datasets. While we observed improvements in certain cases, other results showed minimal changes or slight decreases. We speculate that our model's focus on multiple entities and their relations, coupled with training on images mainly containing multiple objects and their interactions, limits its ability to improve classification tasks that often include only a single object and lack noticeable relations. This discrepancy likely contributes to the mixed results observed in our zero-shot classification experiments.

% To perform classification, we create a template for each class and perform retrieval based on these templates. Similar to the inference process described in Section \ref{subsec:infer}, we use a linear combination of the base VLMs' coarse-grained features and our coarse-grained embeddings to calculate the final similarity score.

\subsection{Compositional Benchmarks}
We use two benchmarks to evaluate the compositional capabilities of our method.
The ARO benchmark \cite{yuksekgonul2022and} is designed to evaluate VLMs' ability to understand various attributes, relationships, and orderings. We utilize two parts of the ARO benchmark: \textit{1) VG-Attribution:} This benchmark involves binary classification tasks where each image is paired with two captions. One caption correctly describes two objects along with their attributes, while the other caption is incorrect because it swaps the attributes of the objects. The models' ability to identify the correct caption is assessed, thereby evaluating their attribute-binding capability. \textit{2) VG-Relation:} Similar to VG-Attribution, this part also consists of binary classification tasks. For each image, there is one correct caption and one incorrect one. The correct caption describes two objects and their relationship, whereas in the incorrect caption, the objects are swapped. This task measures the models' ability to accurately understand relationships and orderings between objects in images.
 
SVO-Probes \cite{hendricks2021probing} is another benchmark designed to evaluate VLMs' understanding of relationships and attributes. The benchmark comprises a dataset of paired images labeled as positive or negative, accompanied by a positive caption and a positive and negative triplet. Each positive caption contains the subject, verb, and object present in its positive triplet, while each negative triplet differs in one of these three parts from the positive triplet. To create a negative caption, we replace positive triplets in the caption with their negative counterpart, which enables the assessment of the model's understanding in both entity recognition (subject, object replacement) and relational understanding (verb replacement) by matching the images with their corresponding positive or negative captions in a binary retrieval task.

We applied our method to CLIP-ViT-B/32, CLIP-ViT-L/14, COCA, and NegCLIP and evaluated their performance on the specified compositional benchmarks. As shown in Table \ref{tb:compositional}, we observe general performance improvements, with only a few exceptions. These enhancements are attributed to the fine-grained components we constructed during training. The entity components enhance the models' ability to bind objects with their attributes, while the relational components improve their understanding of relationships. Notably, the improvement gap is typically higher in the VG-Attribution benchmark compared to the VG-Relation benchmark. This difference may be because VG-Relation also assesses the models' capability to recognize order, which is not addressed by our method.

 % \subsection{Winoground Benchmark}
 % Winoground represents a benchmark and dataset for evaluating compositional reasoning in VLMs. For this task, we present two pairs of images accompanied by two captions. Each pair contains identical objects arranged in distinct relations and orders. The model is expected to discern the differing relations and accurately match each image with its corresponding caption. As demonstrated in Table 3, our method exhibits enhanced proficiency in understanding the relationships between objects when compared to competing models.

\begin{table}
  \caption{Results on compositional benchmarks, including attribute binding (VG-Att), subject-object binding (VG-Rel), and SVO-Probes.}
  \label{tb:compositional}
  \centering
  \setlength{\tabcolsep}{6pt}
  \resizebox{0.5\textwidth}{!}{
  \begin{tabular}{l|lll}
              \multicolumn{1}{c}{\textbf{Method}} & \textbf{VG-Rel} & \textbf{VG-Att} & \begin{tabular}{@{}c@{}} \textbf{SVO} \\ \textbf{Probes} \end{tabular} \\ 
              \midrule
CLIP-ViT-B/32 & 58.82     & 61.05   & 67.63          \\[-0.4ex]
CLIP-ViT-B/32 + ComAlign  & 61.95       & 66.60 & 70.07            \\[-0.4ex]
Relative Gain          & 3.13       & 5.55   & 2.44            \\ \midrule
COCA-ViT-B/32 & 42.30     & 57.80 & 72.47         \\[-0.4ex]
COCA-ViT-B/32 + ComAlign  & 63.46       & 61.36  & 72.60       \\[-0.4ex]
Relative Gain          & 21.16       & 3.56  & 0.13    \\ \midrule
NegClip-ViT-B/32 & 78.61     & 68.98   & 72.41       \\[-0.4ex]
NegClip-ViT-B/32 + ComAlign  & 79.49       & 71.79   & 72.60          \\[-0.4ex]
Relative Gain          & 0.88       & 2.81   & 0.19           \\ \midrule
CLIP-ViT-L/14 & 60.98     & 60.96   & 70.81          \\[-0.4ex]
CLIP-ViT-L/14 + ComAlign  & 59.53       & 65.90   & 74.01         \\[-0.4ex]
Relative Gain          & -1.45       & 4.94  & 3.2          \\
\bottomrule

% PyramidCLIP-ViT-B/32 & 36.05     & 66.91  & 0          \\[-0.5ex]
% PyramidCLIP-ViT-B/32 + Our  & 38.07      & 66.05   & 0           \\[-0.5ex]
% Relative Gain          & 2.02       & -0.86  & 0          \\ 
\end{tabular}
}
\end{table}

\subsection{Ablation Study}

We conducted several experiments to evaluate our model's performance under different hyper-parameters and ablation conditions.
All experiments used CLIP-ViT-B/32 as the base model, trained exclusively on the Visual Genome and Flickr datasets. The results are reported for the image-to-text and text-to-image retrieval tasks on the MSCOCO validation split. Furthermore, only the image transformer was trained.

\textbf{Loss Term Study}
In this experiment, we examine the effect of fine-grained entity, relation, and global similarities by removing them from the final loss calculation. In this part, we only prevent the addition of the similarity term of these parts to the final loss in Equations \ref{eq:6} and \ref{eq:7} while still allowing all three features to attend to each other within the transformer architecture. Additionally, the omitted similarity terms will not be used during inference. The results can be seen in Table \ref{tb:table3}.

In the second part of the experiment, in addition to excluding them from the loss calculation, we prevent the omitted features from interacting with others within the transformer. By doing so, as seen in Table \ref{tb:table4}, we observe a further decrease in performance, compared to only removing the loss term. This suggests that even though fine-grained features are not used during inference, their interaction with global features enhances overall alignment.

\begin{table}[t]
\centering
\caption{Ablation study of each loss term on MSCOCO Zero-Shot Retrieval.}
\label{tb:table3}
\setlength{\tabcolsep}{6pt}
\resizebox{0.5\textwidth}{!}{
\begin{tabular}{llllcc}
\toprule
 \multirow{3}{*}{\textbf{Base VLM}} & \multicolumn{3}{c}{\textbf{Loss Term}} & \multicolumn{2}{c}{\textbf{MSCOCO}} \\
 \cmidrule(lr){2-4} \cmidrule(lr){5-6}
  & Global & Entity & Relation & I2T R@1 & T2I R@1 \\
\midrule  
\multirow{5}{*}{ViT-B/32} & \checkmark & \checkmark & \checkmark & 54.60 & 37.07\\
 & \checkmark & \checkmark &  & 52.24 & 36.50\\
 & \checkmark &  & \checkmark & 52.76 & 35.19\\
 & \checkmark &  &  & 53.76 & 36.25\\
 &  & \checkmark & \checkmark & 52.26 & 36.37\\
 \bottomrule
\end{tabular}
}
\end{table}

\begin{table}[t]
\centering
\caption{Ablation study of each feature on MSCOCO Zero-Shot Retrieval. (Removal of loss term alongside exclusion from the transformer)}
\label{tb:table4}
\setlength{\tabcolsep}{6pt}
\resizebox{0.5\textwidth}{!}{
\begin{tabular}{llllcc}
\toprule
 \multirow{3}{*}{\textbf{Base VLM}} & \multicolumn{3}{c}{\textbf{Feature}} & \multicolumn{2}{c}{\textbf{MSCOCO}} \\
 \cmidrule(lr){2-4} \cmidrule(lr){5-6}
  & Global & Entity & Relation & I2T R@1 & T2I R@1 \\
\midrule  
\multirow{5}{*}{ViT-B/32} & \checkmark & \checkmark & \checkmark & 54.60 & 37.07\\
 & \checkmark & \checkmark &  & 53.24 &  35.83\\
 & \checkmark &  & \checkmark & 52.64 & 35.61\\
 & \checkmark &  &  & 52.10 & 34.14\\
 &  & \checkmark & \checkmark & 52.38 & 35.16\\
 \bottomrule
\end{tabular}
}
\end{table}

\textbf{Network Architecture}
In this section, we examine the effects of different encoder architectures as alternatives to transformer layers. We experimented with two new architectures: one using fully connected layers that process both coarse and fine-grained features through a single network and another using two distinct networks for each feature type. As shown in Table \ref{tb:ablation-architecture}, the transformer-based architecture yields superior results, likely due to its ability to facilitate interactions between different features.

\begin{table}[h]
\centering
\caption{Ablation study of Architecture on MSCOCO Zero-Shot Retrieval.}
\label{tb:ablation-architecture}
\setlength{\tabcolsep}{6pt}
\resizebox{0.4\textwidth}{!}{
\begin{tabular}{ccc}
\toprule
 \multirow{3}{*}{\textbf{Architecture}} & \multicolumn{2}{c}{\textbf{MSCOCO}} \\
 \cmidrule(lr){2-3}
   & I2T R@1 & T2I R@1 \\
   \midrule
   transformer & 54.78 & 37.60 \\
   FC (shared network) & 53.62 & 35.33 \\
   FC (separate networks) & 53.62 & 34.87 \\
 \bottomrule
\end{tabular}
}
\end{table}

\textbf{Network Layers}
 In this experiment, we examined how the size of the appended network affects model performance. Specifically, we increased the number of layers in the transformer network to enhance its expressive power. However, as shown in Table \ref{tb:ablation-layers}, increasing the number of layers to four significantly decreased performance. We believe this is due to the small size of our dataset, which leads to overfitting when using a larger network.

\begin{table}[h]
\centering
\caption{Ablation study of the number of transformer layers on MSCOCO Zero-Shot Retrieval.}
\label{tb:ablation-layers}
\setlength{\tabcolsep}{6pt}
\resizebox{0.4\textwidth}{!}{
\tiny 
\begin{tabular}{ccc}
\midrule
 \multirow{3}{*}{\textbf{Number of Layers}} & \multicolumn{2}{c}{\textbf{MSCOCO}} \\
 \cmidrule(lr){2-3}
   & I2T R@1 & T2I R@1 \\
   \midrule[0.3px]
   1 & 54.78 & 37.60 \\
   2 & 54.60 & 37.07 \\
   4 & 28.94 & 19.71 \\
 \midrule
\end{tabular}
}
\end{table}

\subsection{Visualization}

\begin{figure}[h]
    \centering
    
    \includegraphics[width=\linewidth]{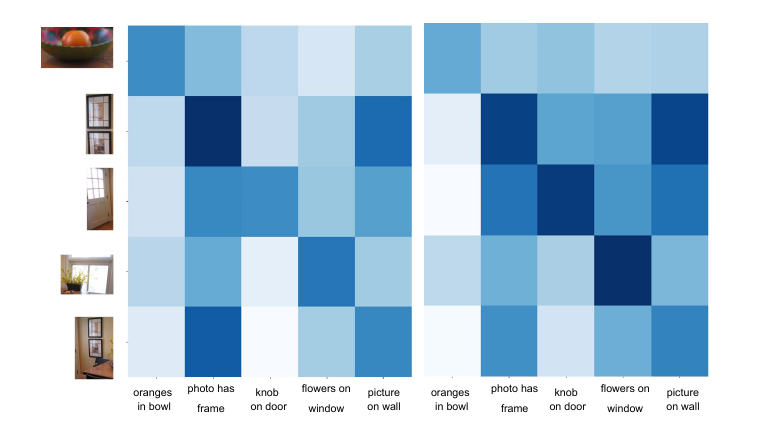}
    \caption{Illustration of relational component similarity matrices. Left: CLIP-ViT-B/32, Right: ComAlign (Ours).}
    \label{relComp}
\end{figure}

\begin{figure}[h]
    \centering
    
    \includegraphics[width=\linewidth]{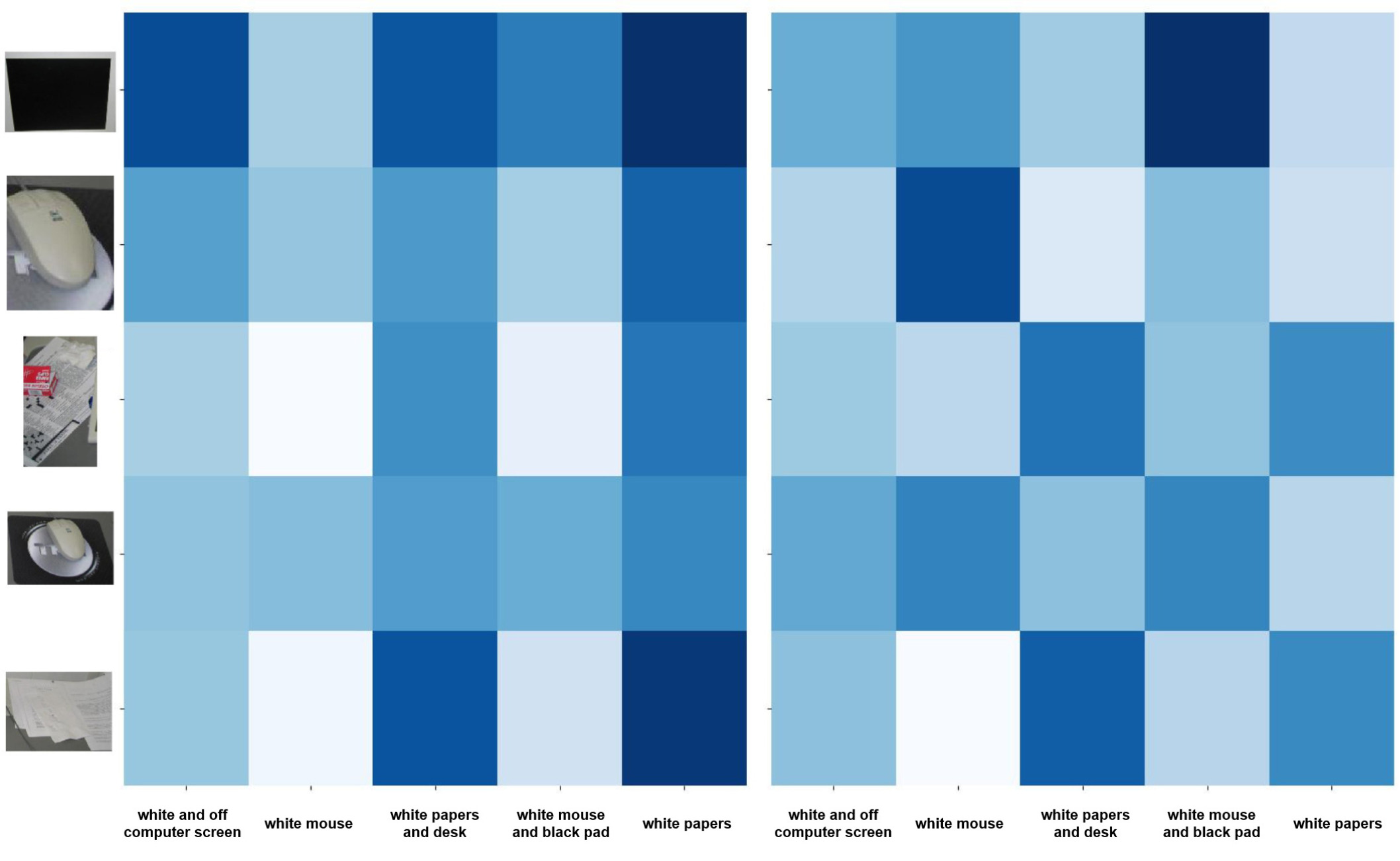}
    \caption{Illustration of entity component similarity matrices. Left: CLIP-ViT-B/32, Right: ComAlign (Ours).}
    \label{entComp}
\end{figure}

As illustrated in the similarity matrix of Figure \ref{relComp} and Figure \ref{entComp}, our alignment surpasses CLIP in matching textual and visual components for both entity and relation. We compute the similarity matrix of five pairs of textual and visual components for relation and entity using CLIP-ViT-B/32 and our own method. Our method exhibits superior performance, as evident by the higher values along the diagonal of the matrix. In addition to the diagonal values, other matrix elements may indicate semantic relevance, and our alignment demonstrates better performance in matching these.

\vspace{-\baselineskip} % Adjust this value to control the spacing between the two figures

\section{Conclusion}
\label{Conclusion}
In this paper, we proposed an alignment model to enhance the compositional understanding of VLMs while maintaining the coarse-grained features. Our approach involves extracting fine-grained entity and relational components and proposing a strategy to match the corresponding components across modalities. We have shown that it is possible to align the base VLMs using a lightweight network and a relatively small dataset to utilize their fine-grained and compositional capacity more efficiently. By enhancing the fine-grained and compositional understanding of VLMs, we improve retrieval, compositional understanding, and downstream tasks.

\textbf{Limitations and Future Works} 
Although our method incorporates elements of text structure, it fails to comprehend the direction of relationships between objects. Furthermore, we do not fully utilize the entire graph structure; instead, we only match nodes and edges of relational components. Future works can involve addressing these limitations to potentially improve performance.

% As mentioned earlier, the methodology employed in this study does not address compositional tasks involving the reordering of objects within captions. Consequently, the performance gains observed on the ARO (VG-Rel) benchmark are relatively more minor. Future research endeavors could concentrate on integrating negative samples to potentially enhance the approach's capability in handling such compositional tasks. Additionally, a limitation of our method is that we create compositional components solely by cropping images, which could contain distracting objects and may be restrictive. Future work could involve generating components using generative models to produce clean training data, potentially enhancing performance.

\newpage
\appendix

\subsection{Hyper-parameter Tuning}
\label{app:Hyper-parameter Tuning}
\begin{figure}
  \centering
  % \fbox{\rule[-.5cm]{0cm}{4cm} \rule[-.5cm]{4cm}{0cm}}
  \includegraphics[width=9.3cm]{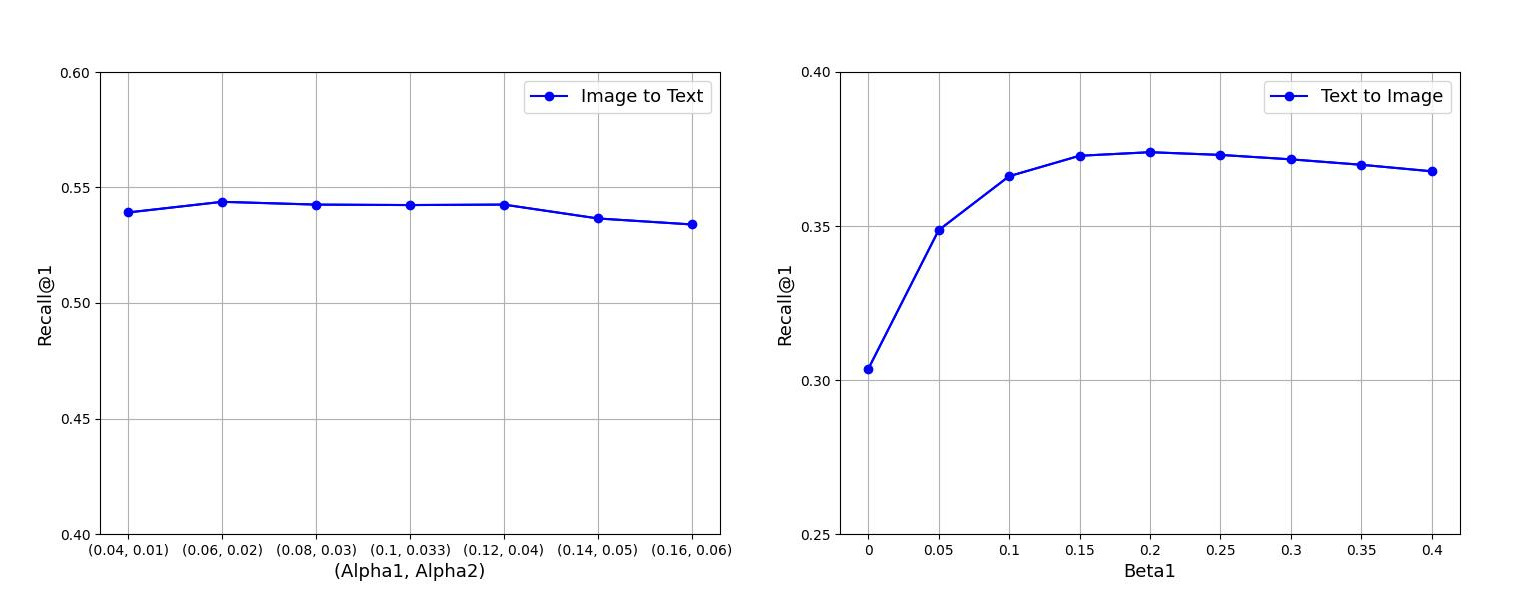}
  \caption{Impact of different values of $\alpha_1$, $\alpha_2$, and $\beta_1$ on I2T and T2I retrieval on our validation set.}
  \label{fig:charts1}
\end{figure}

Hyper-parameters of Equations \ref{eq:10} and \ref{eq:11} have been tuned utilizing a subset of MS-COCO Training split accomplished by ViT-B/32 model which trained on VisualGenome and Flickr30k datasets. Our experimented results are illustrated in Figures \ref{fig:charts1}.

\begin{figure}
  \centering
  % \fbox{\rule[-.5cm]{0cm}{4cm} \rule[-.5cm]{4cm}{0cm}}
  \includegraphics[width=9.3cm]{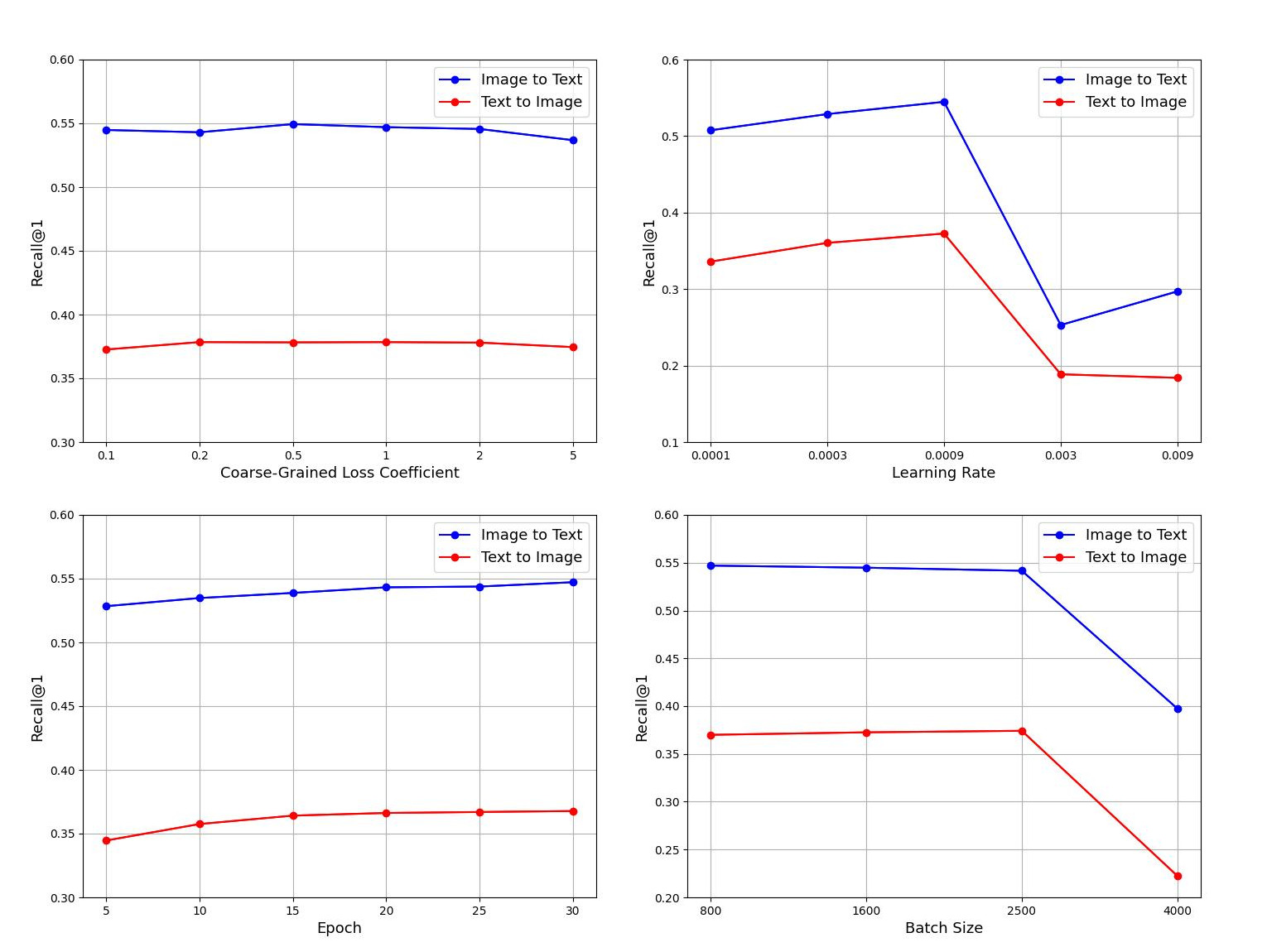}
  \caption{Experiments with different coefficients of coarse-grained contrastive loss, learning rate, number of epochs, and batch sizes on I2T and T2I zero-shot retrieval on MSCOCO.}
  \label{fig:charts3}
\end{figure}

Also, we report the performance of our method under different hyper-parameters in zero-shot image-text retrieval on MSCOCO. Figure \ref{fig:charts3} shows the results of using various batch sizes, learning rates, and training epochs, as well as different coefficients for the coarse-grained contrastive loss.

{\small
\bibliographystyle{splncs04}
\bibliography{references}
}

\end{document}